\documentclass[10pt,twocolumn,a4paper]{article}

\usepackage[utf8]{inputenc}
\usepackage[T1]{fontenc}
\usepackage{lmodern}
\usepackage{orcidlink}

\usepackage[
    a4paper,
    left=2.5cm,
    right=2.5cm,
    top=1.5cm,
    bottom=1.8cm,
    columnsep=0.7cm
]{geometry}

\usepackage{amsmath}
\usepackage{amssymb}
\usepackage{graphicx}
\usepackage{booktabs}
\usepackage{multirow}
\usepackage{microtype}
\usepackage{caption}
\usepackage{xcolor}
\usepackage{listings}
\usepackage{titling}
\usepackage{hyperref}
\usepackage{placeins}
\hypersetup{
    hidelinks
}

\lstdefinestyle{prompt}{
    basicstyle=\ttfamily\small,
    backgroundcolor=\color{gray!8},
    frame=single,
    rulecolor=\color{gray!40},
    breaklines=true,
    breakatwhitespace=false,
    columns=fullflexible,
    keepspaces=true,
    showstringspaces=false,
    numbers=none,
    xleftmargin=0.5em,
    xrightmargin=0.5em,
    aboveskip=1em,
    belowskip=1em
}

\title{
    Cross-Lingual Clinical Entity Projection:
    Comparing Supervised, Hybrid, and Large Language Model Approaches
}

\begin{document}

\twocolumn[
\begin{@twocolumnfalse}

\begin{center}

{\LARGE
Cross-Lingual Clinical Annotation Projection as Constrained Text Generation: A Six-Language Study
\par}

\vspace{0.8em}

{\small
Álvaro Rey-Blanes\textsuperscript{1,2,3,*}\,
\orcidlink{0009-0006-3569-5633}
\hspace{1.4em}
Francisco J. Moreno-Barea\textsuperscript{1,2,3}\,
\orcidlink{0000-0002-3887-9095}
\hspace{1.4em}
Francisco J. Veredas\textsuperscript{1,2,3}\,
\orcidlink{0000-0003-0739-2505}
\par}

\vspace{0.7em}

\begin{minipage}{0.88\textwidth}
\centering
\footnotesize

\textsuperscript{1}Department of Programming Languages and Computer Sciences,
Universidad de Málaga, Málaga, Spain\par

\textsuperscript{2}Research Institute of Multilingual Language Technologies,
Universidad de Málaga, Málaga, Spain\par

\textsuperscript{3}IBIMA Plataforma BIONAND,
Instituto de Investigación Biomédica de Málaga, Málaga, Spain\par

\vspace{0.25em}

\textsuperscript{*}Corresponding author:
\href{mailto:alvaroreyb@uma.es}{alvaroreyb@uma.es}

\end{minipage}

\vspace{1em}

\end{center}

\begin{abstract}

\textbf{Background:}
To determine whether cross-lingual clinical annotation projection can be formulated as a text-preserving, document-level generative task that produces verifiable character-level annotations for multilingual clinical corpus construction, and to characterize its robustness and computational trade-offs relative to candidate-based projection pipelines.

\textbf{Methods:}
We developed a constrained LLM projection workflow that inserts entity tags directly into immutable target-language text, followed by deterministic validation and character-offset reconstruction. We evaluated it alongside supervised candidate-span projection and hybrid ML–LLM refinement for transferring Spanish \textsc{Disease}, \textsc{Symptom}, and \textsc{Procedure} annotations into six languages. Evaluation used MultiClinAI gold standard with strict span matching and character-overlap F1

\textbf{Results:}
Direct LLM projection achieved the strongest and most consistent performance. GLM 5.2 obtained a mean Strict F1 of 0.9201 across 18 language--entity combinations, while locally deployable Gemma4:31B achieved 0.9133. The best LLM configuration improved Strict F1 over the previous state of the art in all 18 settings, by 0.0564–0.1512, yielding 55,416 grounded mentions with reconstructed offsets.

\textbf{Conclusions:}
Direct LLM-based projection enables high-quality multilingual clinical annotation transfer and provides a practical approach for extending clinical NLP resources to languages with fewer annotated datasets and language-specific tools. Combined with local inference and deterministic validation, it can substantially reduce expert time and cost for multilingual clinical corpus construction.
\end{abstract}

\textbf{Keywords}: cross-lingual annotation projection, clinical named entity recognition, multilingual clinical NLP, large language models, clinical corpora

\vspace{5mm}
\end{@twocolumnfalse}
]
\par

\vspace{5mm}

\section{Background and Significance}
\label{sec:background-significance}

Clinical narratives contain essential information on clinical conditions, signs and symptoms, therapeutic interventions, and diagnostic procedures. Automatically identifying these concepts is therefore a central requirement for the secondary use of clinical text in applications such as cohort identification, clinical decision support, epidemiological surveillance, and retrospective research. Named Entity Recognition (NER) provides the span-level representations required by many of these applications, but its development remains highly dependent on manually annotated corpora produced or validated by domain experts~\cite{neveol2018clinical}.

Such resources are unevenly distributed across languages. English benefits from numerous annotated datasets, pretrained models, and evaluation benchmarks, whereas many other languages have substantially fewer reusable clinical corpora. This imbalance is particularly relevant in the clinical domain, where expert annotation is costly and time-consuming.

Cross-lingual annotation projection offers an alternative to independently annotating equivalent resources in every language. Given an annotated source document and its translation, projection methods identify the corresponding target-language span while preserving its semantic label and associated metadata~\cite{yarowsky2001inducing,ehrmann2011building}. A single expert-annotated source corpus can therefore support multiple multilingual resources or provide supervision for target-language extraction systems.

Annotation projection is not equivalent to copying character offsets between texts. Translation may expand, contract, reorder, or reformulate entity mentions, and sentence restructuring can alter their positions. The task therefore requires not only identifying semantic equivalence but also recovering the exact target boundaries required for span-based evaluation and downstream processing.

Clinical language further complicates projection through specialised terminology, abbreviations, numerical expressions, and variable descriptive formulations. Diseases, symptoms, and procedures may also differ in lexical form and annotation granularity across translations. Projection must therefore preserve both conceptual equivalence and exact span boundaries.

Building on these efforts, MultiClinAI provides a common benchmark for multilingual clinical information extraction and cross-lingual corpus construction~\cite{donoso-etal-2026-multiclinai}. It comprises MultiClinNER, which evaluates clinical entity recognition across seven languages, and MultiClinCorpus, which projects \textsc{Disease}, \textsc{Symptom}, and \textsc{Procedure} annotations from Spanish into English, Dutch, Italian, Romanian, Swedish, and Czech.

\subsection{Related Work}
\label{sec:related-work}
Cross-lingual annotation projection transfers annotations from a source document to a semantically equivalent target-language text and has been used to create resources for languages with limited manually labelled data~\cite{yarowsky2001inducing}. Unlike NER, the source entity, boundaries, and semantic category are known; the task is to locate the equivalent target expression and recover its character offsets.

Early approaches relied on statistical word alignment, bilingual dictionaries, machine translation, and string similarity~\cite{ehrmann2011building}. but are less reliable under paraphrasing, reordering, expansion, or contraction. More recent methods use multilingual contextual representations to align tokens or rank candidate spans.

Projected annotations can serve directly as multilingual corpus labels or as weak supervision for target-language NER models~\cite{ni2017weakly}. Direct corpus construction requires each projection to match an exact target substring while preserving entity type and annotation scope.

\subsubsection{Word and Span Alignment}

Word alignment identifies translational correspondences between source and target tokens. Contextual multilingual encoders enable this without relying solely on lexical overlap: SimAlign extracts links without task-specific training~\cite{jalili-sabet-etal-2020-simalign}. whereas AWESoME Align fine-tunes multilingual encoders on parallel data~\cite{dou-neubig-2021-word}.

Token-level correspondences must still be converted into valid entity spans because translations may change the number or continuity of aligned tokens. Proposed solutions include propagating source-span evidence through token alignments~\cite{remy-2026-clinicalaligner}, ranking explicit target-span candidates~\cite{politov-etal-2025-revisiting}. and span-oriented alignment methods designed to preserve annotation boundaries~\cite{jacqmin-etal-2021-spanalign}.

Boundary sensitivity motivates both exact and relaxed biomedical NER evaluation~\cite{tsai-etal-2006-various}. MultiClinAI combines strict entity-type and character-offset matching with character-overlap metrics, distinguishing localisation errors from boundary disagreements.

\subsubsection{Clinical Annotation Projection and MultiClinCorpus}

Clinical annotation projection has been applied through several paradigms. FRASIMED generated French clinical annotations using BERT-based alignment \cite{zaghir-etal-2024-frasimed}; Rodríguez-Miret et al.\ generated Catalan resources from translated Spanish corpora with expert validation \cite{info15100585}; and E3C-3.0 used an LLM-based semi-automatic procedure followed by human revision across several European languages~\cite{Magnini2023, ghosh-etal-2025-low-resource}. These studies demonstrate the feasibility of multilingual clinical corpus construction using encoder-based, translation-based, and generative projection methods.

Within MultiClinAI, the MultiClinCorpus shared task provides a common benchmark for cross-lingual clinical annotation projection~\cite{donoso-etal-2026-multiclinai}. Submitted systems explored markedly different projection paradigms.

The systems submitted to MultiClinCorpus explored three complementary projection paradigms. Team \textit{blue} adopted a lightweight lexical approach based on cognate detection and fuzzy string matching~\cite{sharma-etal-2026-blue-smm4h}. ICB-UMA formulated projection as target-span candidate ranking: candidates were scored using heuristic constraints and ranked by an XGBoost model combining surface, positional, and semantic features, with uncertain predictions optionally revised through LLM-based correction~\cite{rey-blanes-etal-2026-icb} .ClinicalAligner26AM used a domain-adapted cross-lingual token aligner to project source-span evidence and decode target spans, with variants incorporating MultiClinNER predictions and task-specific supervision~\cite{remy-2026-parallia}. These approaches represent lexical matching, supervised span selection with generative refinement, and neural token alignment, and provide the closest methodological comparators for this work.

Despite these advances, conventional projection pipelines remain dependent on explicit alignment, candidate generation, or ranking mechanisms. LLMs offer a document-level alternative that can exploit broader semantic and contextual information without explicit candidate enumeration. However, clinical corpus construction requires projected annotations to remain exactly grounded in the target text, without paraphrasing, normalization, or surface-form alteration, and to yield deterministic character offsets. The key question is therefore whether generative semantic matching can be constrained to preserve textual integrity and produce verifiable, corpus-ready annotations.

This study makes three contributions. First, we formulate projection as constrained document-level text tagging, treating the target document as an immutable textual surface rather than relying on explicit alignment or pre-generated candidates. Second, we couple LLM inference with deterministic validation and character-offset reconstruction to verify tag structure, entity counts, and text preservation and produce reproducible character-level annotations. Third, we evaluate this formulation across six target languages and three entity types against supervised candidate-based and hybrid ML--LLM comparators, characterising robustness, exact-boundary accuracy, and computational trade-offs.

\section{Objective}
\label{sec:objective}

The objective of this study was to determine whether cross-lingual clinical annotation projection can be formulated as a constrained, text-preserving, document-level generative task capable of producing accurate and verifiable character-level annotations for multilingual clinical corpus construction. We further assessed the robustness of this formulation across languages and entity types and characterised its accuracy and computational trade-offs relative to candidate-based supervised and hybrid projection approaches.

\section{Materials and Methods}
\label{sec:methods}

We evaluated three projection strategies: supervised machine learning (ML), hybrid ML--LLM correction, and direct LLM-based projection. All were evaluated on the same official test set using Strict F1 and character-overlap F1. Strict evaluation required exact agreement in entity type, and start and end offsets; unmatched predictions and gold entities were counted as false positives and false negatives, respectively. Precision, recall, and F1 were computed as $P = TP/(TP+FP)$, $R = TP/(TP+FN)$, and $F_1 = 2 \times P \times R/(P+R)$. To quantify partial boundary agreement, overlap similarity between a gold span $g$ and prediction $p$ was defined as $\phi(g,p)=2\times|g\cap p|/(|g|+|p|)$. Character precision ($P_c$) and recall ($R_c$) were obtained from the best overlaps, with $F1_c=2\times P_c\times R_c/(P_c+R_c)$ when $P_c+R_c>0$, and 0 otherwise.

The gold-standard test annotations were hidden from participants; predictions were submitted to the official MultiClinAI evaluation server\footnote{\url{http://temu.bsc.es:8080/}}, which returned the corresponding evaluation scores.

\subsection{Data}
\label{sec:data}

The dataset consists of parallel clinical documents released for the MultiClinCorpus shared task~\cite{multiclinai_corpus, donoso-etal-2026-multiclinai}. Spanish source documents are paired with translations into six target languages: English, Czech, Italian, Dutch, Romanian, and Swedish. The annotations cover three entity types: \textsc{Disease}, \textsc{Procedure}, and \textsc{Symptom}. The corpus is divided into independent training and test partitions, summarised in Table~\ref{tab:corpus-data}.

The complete test partition contains 3,260 parallel documents, with between 59,552 and 81,510 annotations depending on the target language and entity type. Official evaluation was performed on the gold-standard subset reported in Table~\ref{tab:corpus-data}. All test results reported in this work refer exclusively to this evaluated subset.

\begin{table*}[!t]
\centering
\caption{Training and gold-standard test data. Training values indicate the number of annotated entities; test values indicate the number of annotated entities used for gold-standard evaluation. Each target language contains 1,258 training documents.}
\label{tab:corpus-data}
\small
\begin{tabular}{lrrrrrr}
\toprule
& \multicolumn{3}{c}{Training partition}
& \multicolumn{3}{c}{Gold-standard test subset} \\
\cmidrule(lr){2-4}
\cmidrule(lr){5-7}
Target language
& Disease & Procedure & Symptom
& Disease & Procedure & Symptom \\
\midrule
English
& 25,118 & 26,733 & 27,465
&  2,567 &  3,568 &  3,080 \\

Czech
& 25,793 & 27,501 & 27,806
&  2,560 &  3,603 &  3,094 \\

Italian
& 26,159 & 27,394 & 27,929
&  2,569 &  3,553 &  3,087 \\

Dutch
& 25,733 & 27,445 & 27,675
&  2,584 &  3,654 &  3,096 \\

Romanian
& 25,561 & 27,313 & 27,015
&  2,576 &  3,572 &  3,080 \\

Swedish
& 25,580 & 27,079 & 27,531
&  2,555 &  3,587 &  3,095 \\
\bottomrule
\end{tabular}
\end{table*}

\begin{figure*}[!t]
    \centering
    \includegraphics[width=\textwidth]{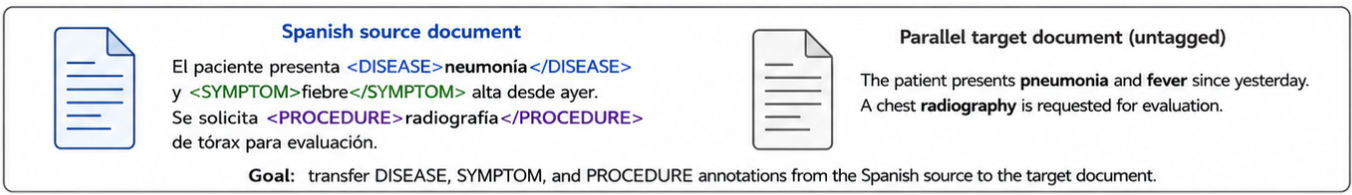}
    \caption{Overview of the main clinical entity cross-lingual projection task.}
    \label{fig:projection-methods}
\end{figure*}

To investigate complementary approaches to cross-lingual clinical entity projection, we evaluated three strategies that differ in the level at which the target span is identified and in the role assigned to the LLM. Figure~\ref{fig:projection-methods} provides an overview of the task to be performed by the three pipelines, which are described in detail in the following subsections.

\subsection{Method 1: Window-based Projection with Machine Learning}
\label{sec:ml}

\begin{figure}[!t]
    \centering
    \includegraphics[width=\linewidth]{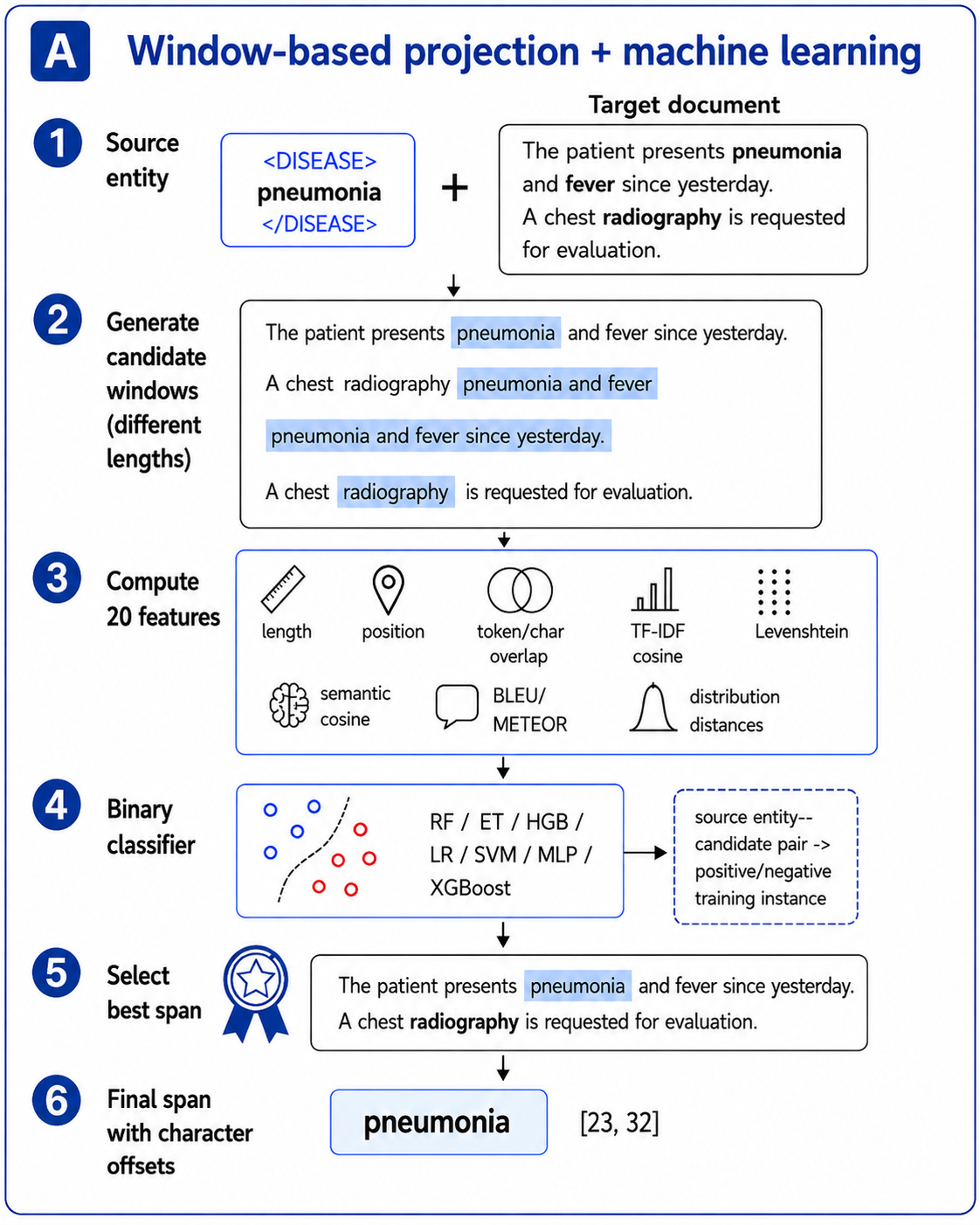}
    \caption{Overview of the window-based machine-learning approach for cross-lingual clinical entity projection. Candidate spans are generated in the target document, represented through surface, positional, structural, and semantic features, and ranked by a supervised binary classifier to select the final projected span.}
    \label{fig:method1}
\end{figure}

We formulated cross-lingual clinical entity projection as a binary classification problem over candidate spans in the target document.  Given an annotated source entity and its parallel target document, candidate windows with lengths close to the source entity were generated. During training, the aligned target entity constituted the positive instance, while nearby and sampled windows constituted negative instances (Figure~\ref{fig:method1}).

Candidate generation combined local hard negatives surrounding the aligned target entity with soft negatives sampled from the target document. Each source entity--candidate pair was represented by 20 features covering length, positional differences, token and character similarity, word and character $n$-gram overlap, term frequency--inverse document frequency (TF--IDF) cosine similarity, Levenshtein and semantic similarity, Jensen--Shannon and Hellinger distances, BLEU, and METEOR~\cite{salton1988term,papineni2002bleu,banerjee2005meteor}. Absolute character positions, document identifiers, and raw text were excluded from the feature matrix; class labels were used only as supervision. 

For each source--target language pair and entity type, source mentions were partitioned into training (80\%) and validation (20\%) subsets. All candidate windows derived from the same mention were assigned to the same subset to prevent information leakage. Each classifier was trained on the training subset, with its decision threshold selected by maximising validation F1.

We compared seven classifiers, sharing training and evaluation protocol: Random Forest, Extra Trees, Histogram Gradient Boosting, class-balanced logistic regression, a support vector machine (SVM) with a radial basis function (RBF) kernel, a multilayer perceptron, and XGBoost~\cite{breiman2001random,geurts2006extremely,friedman2001greedy,chen2016xgboost,cortes1995support,kingma2015adam}. Random Forest and Extra Trees used 500 trees with a minimum leaf size of two; Histogram Gradient Boosting used depth 6, a 0.05 learning rate, and 300 iterations; and the RBF SVM used standardised features with $C=1.0$. The multilayer perceptron used 128- and 64-unit hidden layers, ReLU activations, Adam optimisation, and early stopping. XGBoost used 500 trees, depth 4, a 0.05 learning rate, and row and column subsampling of 0.9.

\subsection{Method 2: Hybrid ML--LLM Refinement}
\label{sec:hybrid}

\begin{figure}[!t]
    \centering
    \includegraphics[width=\linewidth]{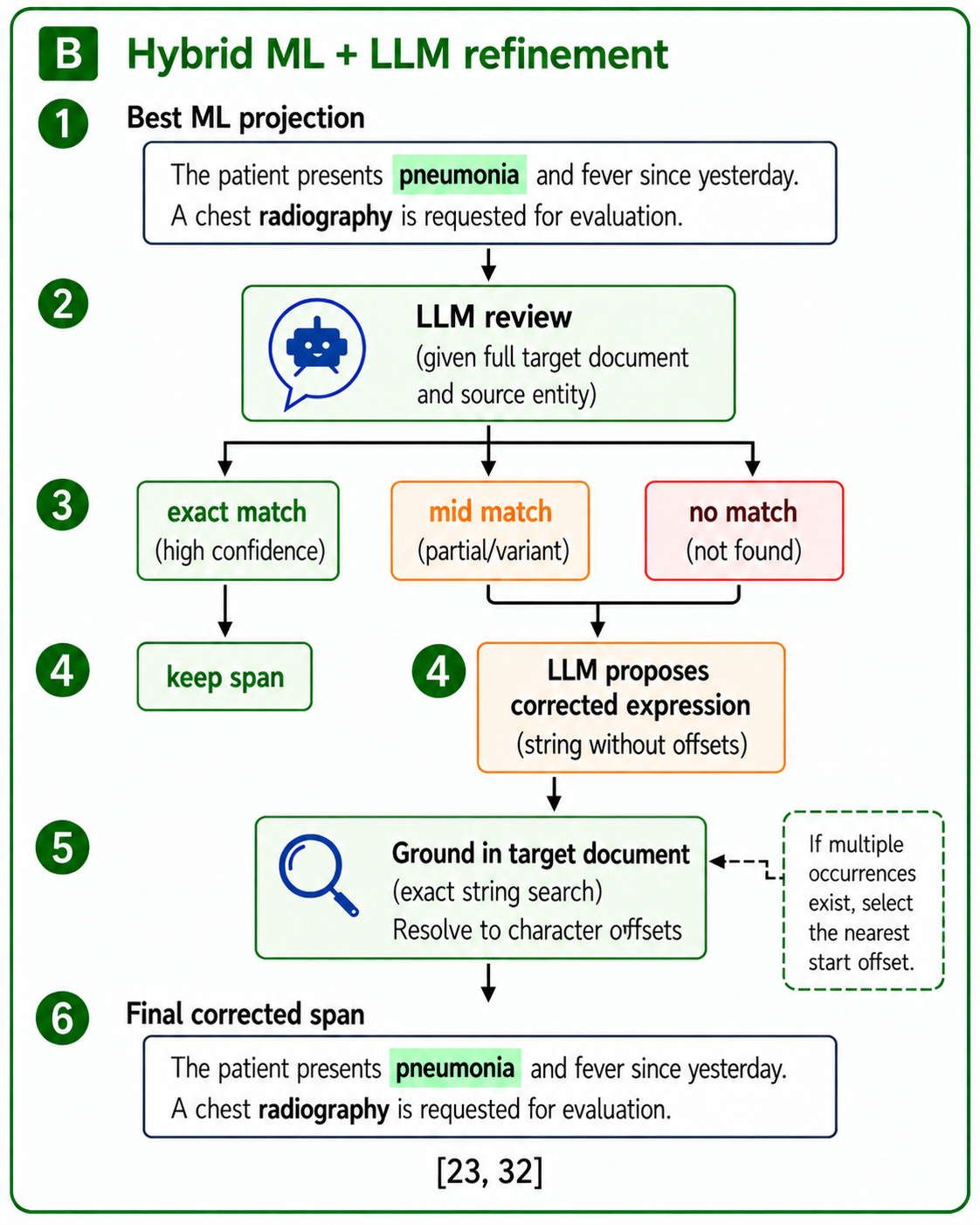}
    \caption{Overview of the hybrid ML--LLM projection approach. The best machine-learning projection is first assessed by an LLM; exact matches are retained, whereas uncertain or incorrect projections are corrected and subsequently grounded in the original target document to obtain valid character offsets.}
    \vspace{1mm}
    \label{fig:method2}
\end{figure}

The hybrid strategy used the ML projection as input and applied an LLM-based refinement step only to potentially incorrect projections. Figure~\ref{fig:method2} summarises this selective correction pipeline, including LLM-based assessment, span correction, and grounding of the resulting expression in the target document. For each target language and entity type, the classifier achieving the highest Strict F1 returned by the official MultiClinAI evaluation \footnote{\url{http://temu.bsc.es:8080/}} was selected as the input to the LLM correction stage. Gold-standard test annotations were not accessible to the participants.

Following the prompt-defined decision criteria, the LLM classified each projection as an \textit{exact match}, \textit{mid match}, or \textit{no match}. Exact matches were retained, whereas \textit{mid match} and \textit{no match} cases were refined using the complete target document and affected projections (see Supplementary Appendix \ref{app:promps} for prompts). The proposed expression was grounded to character offsets by exact string matching. Unique occurrences were accepted directly; when multiple matches existed, the occurrence with the start offset closest to the corresponding Spanish annotation was selected, exploiting the approximate preservation of document structure across translations. Unresolved suggestions were discarded.

\subsection{Method 3: LLM Entity Projection}
\label{sec:llm}

\begin{figure}[!t]
    \centering
    \includegraphics[width=\linewidth]{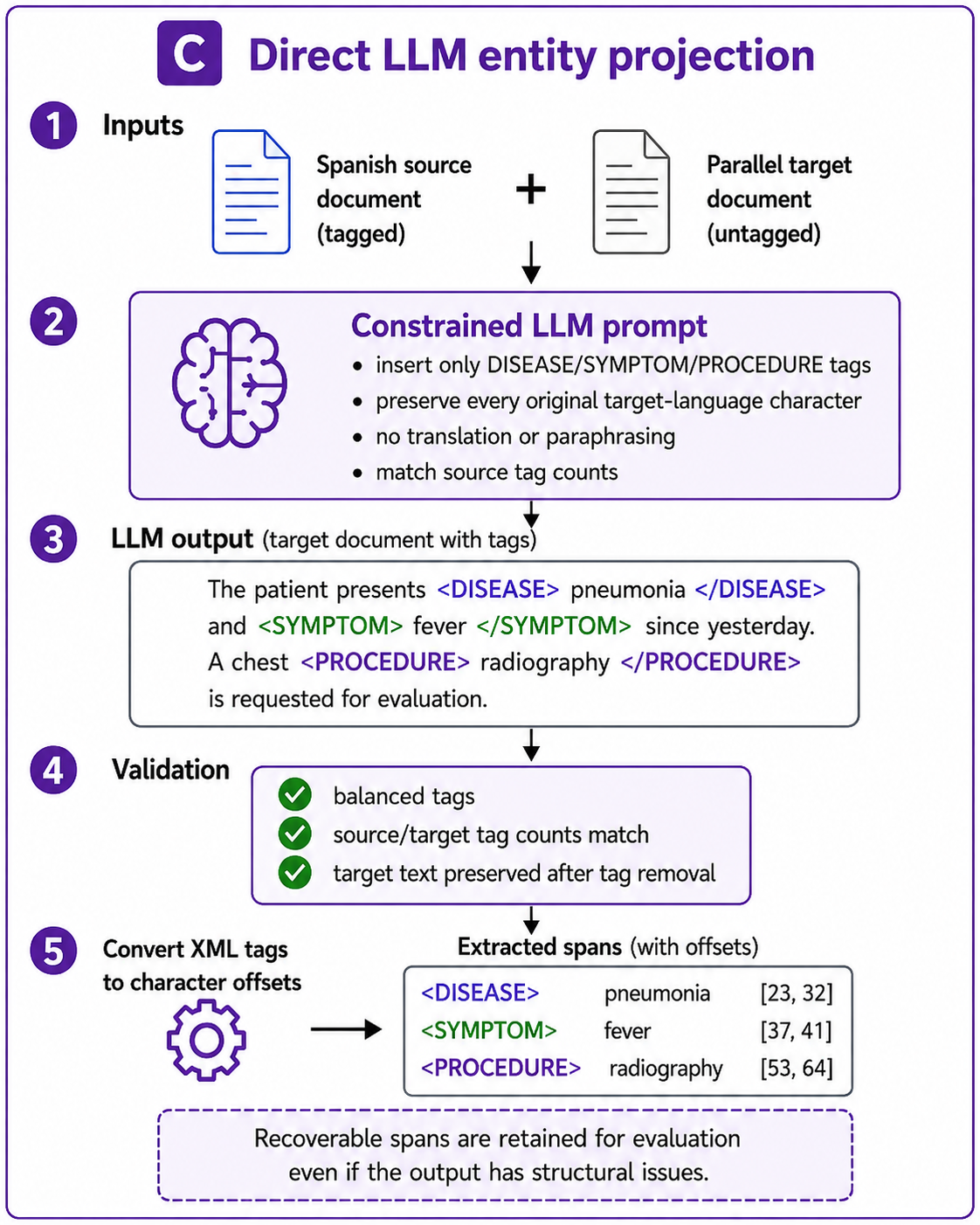}
    \caption{Overview of the direct LLM-based entity projection approach. The tagged Spanish source document and its unannotated parallel translation are provided to the LLM under constrained generation instructions. The resulting tagged target text is validated and converted into character-offset entity spans for evaluation.}
    \label{fig:method3}
\end{figure}

The third approach formulated annotation transfer as a constrained LLM-based span-projection task. Figure~\ref{fig:method3} summarises the pipeline from tagged Spanish source and unannotated target documents to generation, validation, and character-offset reconstruction. The model inserted \texttt{DISEASE}, \texttt{SYMPTOM}, and \texttt{PROCEDURE} tags into the target text while preserving every original character. Source tags were normalised to this label set, and the source annotations determined which entity types were projected for each document pair.

For each document pair, the model received the tagged Spanish source and untagged target document. The prompt (Supplementary Appendix \ref{app:promps}) specified the active labels and required the model to return only the tagged target text, prohibiting translation, paraphrasing, commentary, or any modification beyond XML tag insertion. Tag counts for each active label were required to match the source annotations.

Generation and validation were treated separately. Outputs were checked for balanced tags, matching source--target entity counts, and preservation of the target text after tag removal, using canonicalised line endings and HTML entities. Invalid outputs were retained as \textit{.invalid.txt} sidecars for auditability, while any recoverable mentions resolvable to valid target spans remained included in the quantitative evaluation.

Predictions were converted from inline XML to character offsets by sequentially parsing the output, removing clinical tags, and tracking cumulative character position. This directly yielded start and end offsets, including for repeated expressions, without separate occurrence selection. Strict evaluation required exact agreement in document, entity type, and offsets, while character-overlap F1 captured partial boundary agreement.

\subsection{Models and Inference Settings}
\label{sec:models}

We evaluated three models: \texttt{gemma4:31b}, \texttt{qwen3.6:35b}, and \texttt{glm-5.2}~\cite{gemmateam2026gemma4technicalreport,qwen36_35b,glm5team2026glm5vibecodingagentic}. The Gemma and Qwen models were served locally through an internal server, whereas GLM-5.2 was accessed through the Ollama cloud service. All models received the same source--target document pairs, prompt template, entity-label constraints, and post-processing procedure.

For the local runs, maximum generation length was set to \texttt{num\_predict}=4096 tokens, and the context window was set to \texttt{num\_ctx}=8192. Streaming was enabled, \texttt{think} was set to \texttt{false}, and \texttt{keep\_alive} was set to \texttt{30m}. The connection timeout was set to 30\,s and \texttt{entity\_count\_max\_attempts} was set to 5. Temperature was set to zero.

For the GLM-5.2 cloud runs, streaming was disabled and temperature was set to zero. The request timeout was set to 600\,s, up to three API-level attempts were permitted per document, and a 0.2\,s delay was applied between requests. No explicit \texttt{num\_ctx}, \texttt{num\_predict}, \texttt{keep\_alive}, or \texttt{think} parameter was supplied to the cloud endpoint; these settings therefore remained under the provider defaults.

\section{Results}
\label{sec:results}

Results compare candidate-based projection with generative refinement, direct document-level projection across languages and entity types, exact-boundary performance against previous methods, and the associated computational trade-offs. Previous state-of-the-art results are used as an external reference for assessing the magnitude and consistency of the observed improvements.

\subsection{Candidate-based Projection and the Effect of Generative Refinement}
\label{sec:results-hybrid}

\begin{table*}[!t]
\centering
\caption{Best ML model and LLM refinement by target language and entity type. The ML model is selected by the highest test Strict F1 among the evaluated classifiers. Bold indicates the larger Strict F1 within each ML--LLM pair. Dashes indicate that LLM refinement was not completed.}
\label{tab:ml-llm-results}
\small
\begin{tabular}{lll l rrrrrrr}
\toprule
Lang & Entity & ML Model & System & \multicolumn{6}{c}{Strict} & Char F1 \\
\cmidrule(lr){5-10}
& & & & P & R & F1 & TP & FP & FN & \\
\midrule
\multirow{6}{*}{EN} & \multirow{2}{*}{Disease} & \multirow{2}{*}{Extra Trees} & ML model & 0.7222 & 0.7242 & 0.7232 & 1859 & 715 & 708 & 0.8529 \\
\cline{4-11}
& & & LLM Ref. & 0.8899 & 0.8878 & \textbf{0.8888} & 2279 & 282 & 288 & 0.9522 \\
& \multirow{2}{*}{Procedure} & \multirow{2}{*}{Random Forest} & ML model & 0.6561 & 0.6508 & 0.6534 & 2322 & 1217 & 1246 & 0.7966 \\
\cline{4-11}
& & & LLM Ref. & 0.8343 & 0.8299 & \textbf{0.8321} & 2961 & 588 & 607 & 0.9259 \\
& \multirow{2}{*}{Symptom} & \multirow{2}{*}{Extra Trees} & ML model & 0.6367 & 0.6321 & 0.6344 & 1947 & 1111 & 1133 & 0.8007 \\
\cline{4-11}
& & & LLM Ref. & 0.8759 & 0.8734 & \textbf{0.8747} & 2690 & 381 & 390 & 0.9561 \\
\midrule
\multirow{6}{*}{CS} & \multirow{2}{*}{Disease} & \multirow{2}{*}{Extra Trees} & ML model & 0.4480 & 0.4461 & 0.4471 & 1142 & 1407 & 1418 & 0.5984 \\
\cline{4-11}
& & & LLM Ref. & 0.7982 & 0.7836 & \textbf{0.7909} & 2006 & 507 & 554 & 0.8912 \\
& \multirow{2}{*}{Procedure} & \multirow{2}{*}{Extra Trees} & ML model & 0.4539 & 0.4294 & 0.4413 & 1547 & 1861 & 2056 & 0.5749 \\
\cline{4-11}
& & & LLM Ref. & 0.7587 & 0.7286 & \textbf{0.7433} & 2625 & 835 & 978 & 0.8477 \\
& \multirow{2}{*}{Symptom} & \multirow{2}{*}{Extra Trees} & ML model & 0.3736 & 0.3681 & 0.3708 & 1139 & 1910 & 1955 & 0.5605 \\
\cline{4-11}
& & & LLM Ref. & 0.7521 & 0.7414 & \textbf{0.7467} & 2294 & 756 & 800 & 0.8754 \\
\midrule
\multirow{6}{*}{IT} & \multirow{2}{*}{Disease} & \multirow{2}{*}{Extra Trees} & ML model & 0.7086 & 0.7088 & 0.7087 & 1821 & 749 & 748 & 0.7571 \\
\cline{4-11}
& & & LLM Ref. & 0.8993 & 0.8898 & \textbf{0.8945} & 2286 & 256 & 283 & 0.9503 \\
& \multirow{2}{*}{Procedure} & \multirow{2}{*}{Extra Trees} & ML model & 0.6630 & 0.6583 & 0.6606 & 2339 & 1189 & 1214 & 0.7271 \\
\cline{4-11}
& & & LLM Ref. & 0.8573 & 0.8486 & \textbf{0.8529} & 3015 & 502 & 538 & 0.9249 \\
& \multirow{2}{*}{Symptom} & \multirow{2}{*}{Logistic Regression} & ML model & 0.6310 & 0.6093 & 0.6200 & 1881 & 1100 & 1206 & 0.7980 \\
\cline{4-11}
& & & LLM Ref. & 0.8259 & 0.8024 & \textbf{0.8140} & 2477 & 522 & 610 & 0.9319 \\
\midrule
\multirow{6}{*}{NL} & \multirow{2}{*}{Disease} & \multirow{2}{*}{Extra Trees} & ML model & 0.5748 & 0.5712 & 0.5730 & 1476 & 1092 & 1108 & 0.6671 \\
\cline{4-11}
& & & LLM Ref. & 0.8008 & 0.7810 & \textbf{0.7908} & 2018 & 502 & 566 & 0.8853 \\
& \multirow{2}{*}{Procedure} & \multirow{2}{*}{Extra Trees} & ML model & 0.5618 & 0.5290 & 0.5449 & 1933 & 1508 & 1721 & 0.6431 \\
\cline{4-11}
& & & LLM Ref. & 0.7681 & 0.7351 & \textbf{0.7512} & 2686 & 811 & 968 & 0.8592 \\
& \multirow{2}{*}{Symptom} & \multirow{2}{*}{Extra Trees} & ML model & 0.4839 & 0.4761 & 0.4800 & 1473 & 1571 & 1621 & 0.6194 \\
\cline{4-11}
& & & LLM Ref. & 0.7567 & 0.7489 & \textbf{0.7528} & 2317 & 745 & 777 & 0.8911 \\
\midrule
\multirow{6}{*}{RO} & \multirow{2}{*}{Disease} & \multirow{2}{*}{Extra Trees} & ML model & 0.6527 & 0.6522 & 0.6524 & 1680 & 894 & 896 & 0.7313 \\
\cline{4-11}
& & & LLM Ref. & 0.8659 & 0.8521 & \textbf{0.8589} & 2195 & 340 & 381 & 0.9375 \\
& \multirow{2}{*}{Procedure} & \multirow{2}{*}{Extra Trees} & ML model & 0.6119 & 0.5971 & 0.6044 & 2133 & 1353 & 1439 & 0.7072 \\
\cline{4-11}
& & & LLM Ref. & 0.8472 & 0.8287 & \textbf{0.8378} & 2960 & 534 & 612 & 0.9274 \\
& \multirow{2}{*}{Symptom} & \multirow{2}{*}{Extra Trees} & ML model & 0.5084 & 0.5026 & 0.5055 & 1548 & 1497 & 1532 & 0.6155 \\
\cline{4-11}
& & & LLM Ref. & 0.8021 & 0.7987 & \textbf{0.8004} & 2460 & 607 & 620 & 0.9150 \\
\midrule
\multirow{6}{*}{SV} & \multirow{2}{*}{Disease} & \multirow{2}{*}{Random Forest} & ML model & 0.5439 & 0.5436 & 0.5437 & 1389 & 1165 & 1166 & 0.6635 \\
\cline{4-11}
& & & LLM Ref. & 0.8002 & 0.7933 & \textbf{0.7968} & 2027 & 506 & 528 & 0.8918 \\
& \multirow{2}{*}{Procedure} & \multirow{2}{*}{Extra Trees} & ML model & 0.5343 & 0.5040 & 0.5187 & 1808 & 1576 & 1779 & 0.6362 \\
\cline{4-11}
& & & LLM Ref. & 0.7807 & 0.7513 & \textbf{0.7657} & 2695 & 757 & 892 & 0.8749 \\
& \multirow{2}{*}{Symptom} & \multirow{2}{*}{Extra Trees} & ML model & 0.4672 & 0.4582 & 0.4626 & 1418 & 1617 & 1677 & 0.6309 \\
\cline{4-11}
& & & LLM Ref. & 0.7900 & 0.7780 & \textbf{0.7840} & 2408 & 640 & 687 & 0.9184 \\
\bottomrule
\end{tabular}

\end{table*}

Candidate-based projection showed substantial variation across languages and entity types (Table~\ref{tab:ml-llm-results}). Tree-based ensembles provided the strongest ML baselines, yielding the highest Strict F1 in 17 of the 18 language--entity settings: Extra Trees was selected in 15 cases, Random Forest in two, and logistic regression in one. Performance was highest for English and Italian and lowest for Czech, while \textsc{Disease} consistently achieved higher Strict F1 than \textsc{Procedure} and \textsc{Symptom}. Mean Strict F1 for \textsc{Symptom} was 0.5122, compared with 0.5706 for \textsc{Procedure}. These differences indicate that candidate-based projection is sensitive to the linguistic and boundary characteristics of the target expression despite the inclusion of multilingual semantic features.

Generative refinement substantially reduced this dependency on the initial ML decision. In every completed comparison, LLM refinement increased Strict F1, with a mean absolute gain of 0.2462, while reducing both false-positive and false-negative projections. Character-overlap F1 increased from a mean of 0.6878 for the ML baselines to 0.9087 after refinement, corresponding to a mean absolute gain of 0.2209. The larger character-overlap scores relative to Strict F1 indicate that a substantial proportion of the residual errors involved inaccurate span boundaries rather than complete failure to localise the target entity. Nevertheless, because refinement operates on an upstream candidate-based projection, its ability to recover an entity remains dependent on the information supplied by the preceding projection stage.

\subsection{Robustness of Document-level Projection across Languages and Entity Types}
\label{sec:results-llm}

\begin{table*}[!t]
\centering
\begingroup

\captionsetup{
    width=\linewidth,
    justification=justified,
    singlelinecheck=false,
    skip=10pt
}

\caption{Strict span-level results and character-overlap F1 by target language and entity type. Bold indicates the highest Strict F1 among the available models within each pair. }
\label{tab:llm-projection-results}

\scriptsize
\setlength{\tabcolsep}{3pt}
\renewcommand{\arraystretch}{1.06}

\begin{tabular}{@{}lllrrrrrrr@{}}
\toprule
Lang & Entity & Model & \multicolumn{6}{c}{Strict} & Char \\
\cmidrule(lr){4-9}
     &        &       & P & R & F1 & TP & FP & FN & F1 \\
\midrule

\multirow{9}{*}{EN}
& \multirow{3}{*}{Disease}
& Gemma4:31B  & 0.9590 & 0.9657 & 0.9623 & 2479 & 106 & 88  & 0.9868 \\
& & Qwen3.6:35B& 0.8915 & 0.8703 & 0.8807 & 2234 & 272 & 333  & 0.9493 \\
& & GLM 5.2  & 0.9684 & 0.9677 & \textbf{0.9680} & 2484 & 81 & 83  & 0.9851 \\
\cmidrule(lr){2-10}
& \multirow{3}{*}{Procedure}
& Gemma4:31B  & 0.9294 & 0.9302 & 0.9298 & 3319 & 252 & 249  & 0.9780 \\
& & Qwen3.6:35B& 0.8947 & 0.8307 & 0.8615 & 2964 & 349 & 604  & 0.9303 \\
& & GLM 5.2  & 0.9401 & 0.9372 & \textbf{0.9387} & 3344 & 213 & 224  & 0.9786 \\
\cmidrule(lr){2-10}
& \multirow{3}{*}{Symptom}
& Gemma4:31B  & 0.9445 & 0.9445 & 0.9445 & 2909 & 171 & 171  & 0.9826 \\
& & Qwen3.6:35B& 0.8683 & 0.8390 & 0.8534 & 2584 & 392 & 496  & 0.9396 \\
& & GLM 5.2  & 0.9599 & 0.9471 & \textbf{0.9534} & 2917 & 122 & 163  & 0.9795 \\

\midrule

\multirow{9}{*}{CS}
& \multirow{3}{*}{Disease}
& Gemma4:31B  & 0.9135 & 0.9199 & 0.9167 & 2355 & 223 & 205  & 0.9719 \\
& & Qwen3.6:35B& 0.8571 & 0.8270 & 0.8417 & 2117 & 353 & 443  & 0.9275 \\
& & GLM 5.2  & 0.9250 & 0.9305 & \textbf{0.9278} & 2382 & 193 & 178  & 0.9740 \\
\cmidrule(lr){2-10}
& \multirow{3}{*}{Procedure}
& Gemma4:31B  & 0.8896 & 0.8837 & 0.8867 & 3184 & 395 & 419  & 0.9645 \\
& & Qwen3.6:35B& 0.8627 & 0.7880 & 0.8236 & 2839 & 452 & 764  & 0.9135 \\
& & GLM 5.2  & 0.8979 & 0.8912 & \textbf{0.8946} & 3211 & 365 & 392  & 0.9646 \\
\cmidrule(lr){2-10}
& \multirow{3}{*}{Symptom}
& Gemma4:31B  & 0.9038 & 0.8992 & 0.9015 & 2782 & 296 & 312  & 0.9710 \\
& & Qwen3.6:35B& 0.8427 & 0.7999 & 0.8208 & 2475 & 462 & 619  & 0.9283 \\
& & GLM 5.2  & 0.9196 & 0.9131 & \textbf{0.9163} & 2825 & 247 & 269  & 0.9752 \\

\midrule

\multirow{9}{*}{IT}
& \multirow{3}{*}{Disease}
& Gemma4:31B  & 0.9656 & 0.9720 & \textbf{0.9688} & 2497 & 89 & 72  & 0.9874 \\
& & Qwen3.6:35B& 0.9025 & 0.8505 & 0.8758 & 2185 & 236 & 384  & 0.9397 \\
& & GLM 5.2  & 0.9670 & 0.9685 & 0.9677 & 2488 & 85 & 81  & 0.9851 \\
\cmidrule(lr){2-10}
& \multirow{3}{*}{Procedure}
& Gemma4:31B  & 0.9451 & 0.9552 & \textbf{0.9502} & 3394 & 197 & 159  & 0.9833 \\
& & Qwen3.6:35B& 0.9208 & 0.8283 & 0.8721 & 2943 & 253 & 610  & 0.9214 \\
& & GLM 5.2  & 0.9430 & 0.9538 & 0.9484 & 3389 & 205 & 164  & 0.9851 \\
\cmidrule(lr){2-10}
& \multirow{3}{*}{Symptom}
& Gemma4:31B  & 0.9203 & 0.9206 & 0.9205 & 2842 & 246 & 245  & 0.9860 \\
& & Qwen3.6:35B& 0.8044 & 0.7752 & 0.7895 & 2393 & 582 & 694  & 0.9354 \\
& & GLM 5.2  & 0.9391 & 0.9397 & \textbf{0.9394} & 2901 & 188 & 186  & 0.9900 \\

\midrule

\multirow{9}{*}{NL}
& \multirow{3}{*}{Disease}
& Gemma4:31B  & 0.8787 & 0.8777 & 0.8782 & 2268 & 313 & 316  & 0.9417 \\
& & Qwen3.6:35B& 0.8282 & 0.7910 & 0.8092 & 2044 & 424 & 540  & 0.9018 \\
& & GLM 5.2  & 0.8949 & 0.8932 & \textbf{0.8941} & 2308 & 271 & 276  & 0.9463 \\
\cmidrule(lr){2-10}
& \multirow{3}{*}{Procedure}
& Gemma4:31B  & 0.8579 & 0.8407 & 0.8492 & 3072 & 509 & 582  & 0.9340 \\
& & Qwen3.6:35B& 0.8216 & 0.7411 & 0.7793 & 2708 & 588 & 946  & 0.8805 \\
& & GLM 5.2  & 0.8640 & 0.8487 & \textbf{0.8563} & 3101 & 488 & 553  & 0.9371 \\
\cmidrule(lr){2-10}
& \multirow{3}{*}{Symptom}
& Gemma4:31B  & 0.8343 & 0.8311 & \textbf{0.8327} & 2573 & 511 & 523  & 0.9335 \\
& & Qwen3.6:35B& 0.7664 & 0.7258 & 0.7455 & 2247 & 685 & 849  & 0.8862 \\
& & GLM 5.2  & 0.8315 & 0.8224 & 0.8269 & 2546 & 516 & 550  & 0.9318 \\

\midrule

\multirow{9}{*}{RO}
& \multirow{3}{*}{Disease}
& Gemma4:31B  & 0.9532 & 0.9557 & 0.9544 & 2462 & 121 & 114  & 0.9854 \\
& & Qwen3.6:35B& 0.9019 & 0.8498 & 0.8751 & 2189 & 238 & 387  & 0.9406 \\
& & GLM 5.2  & 0.9560 & 0.9604 & \textbf{0.9582} & 2474 & 114 & 102  & 0.9877 \\
\cmidrule(lr){2-10}
& \multirow{3}{*}{Procedure}
& Gemma4:31B  & 0.9420 & 0.9418 & 0.9419 & 3364 & 207 & 208  & 0.9827 \\
& & Qwen3.6:35B& 0.9152 & 0.8099 & 0.8593 & 2893 & 268 & 679  & 0.9120 \\
& & GLM 5.2  & 0.9448 & 0.9490 & \textbf{0.9469} & 3390 & 198 & 182  & 0.9859 \\
\cmidrule(lr){2-10}
& \multirow{3}{*}{Symptom}
& Gemma4:31B  & 0.9240 & 0.9234 & 0.9237 & 2844 & 234 & 236  & 0.9831 \\
& & Qwen3.6:35B& 0.8324 & 0.8016 & 0.8167 & 2469 & 497 & 611  & 0.9369 \\
& & GLM 5.2  & 0.9487 & 0.9484 & \textbf{0.9485} & 2921 & 158 & 159  & 0.9868 \\

\midrule

\multirow{9}{*}{SV}
& \multirow{3}{*}{Disease}
& Gemma4:31B  & 0.8913 & 0.9014 & 0.8963 & 2303 & 281 & 252  & 0.9626 \\
& & Qwen3.6:35B& 0.8283 & 0.8082 & 0.8181 & 2065 & 428 & 490  & 0.9228 \\
& & GLM 5.2  & 0.9031 & 0.9151 & \textbf{0.9090} & 2338 & 251 & 217  & 0.9676 \\
\cmidrule(lr){2-10}
& \multirow{3}{*}{Procedure}
& Gemma4:31B  & 0.8935 & 0.8938 & \textbf{0.8937} & 3206 & 382 & 381  & 0.9697 \\
& & Qwen3.6:35B& 0.8539 & 0.7775 & 0.8140 & 2789 & 477 & 798  & 0.9124 \\
& & GLM 5.2  & 0.8948 & 0.8918 & 0.8933 & 3199 & 376 & 388  & 0.9694 \\
\cmidrule(lr){2-10}
& \multirow{3}{*}{Symptom}
& Gemma4:31B  & 0.8903 & 0.8866 & \textbf{0.8885} & 2744 & 338 & 351  & 0.9730 \\
& & Qwen3.6:35B& 0.8262 & 0.7819 & 0.8035 & 2420 & 509 & 675  & 0.9206 \\
& & GLM 5.2  & 0.8801 & 0.8682 & 0.8741 & 2687 & 366 & 408  & 0.9650 \\
\bottomrule
\end{tabular}

\endgroup
\end{table*}

Direct document-level projection achieved high Strict F1 across all six target languages and three entity types (Table~\ref{tab:llm-projection-results}). GLM 5.2 obtained the highest overall mean Strict F1 (0.9201), followed closely by Gemma4:31B (0.9133), while Qwen3.6:35B obtained 0.8300; GLM 5.2 led in 13 of 18 language--entity combinations and Gemma4:31B in five. English, Italian, and Romanian produced the strongest overall results, whereas Dutch was the most challenging target language. \textsc{Disease} was the highest-performing entity type for all three models, followed by \textsc{Procedure} and \textsc{Symptom}.

Character-overlap F1 was consistently higher than Strict F1, with macro-averages of 0.9719 for GLM 5.2, 0.9710 for Gemma4:31B, and 0.9222 for Qwen3.6:35B. This indicates that many residual errors involved entity-boundary mismatches rather than localisation of an unrelated span.

\subsection{Exact-boundary Recovery and Improvement over Previous Projection Methods}
\label{sec:results-sota}
\begin{table*}[!t]
\centering
\caption{Exact-span projection performance relative to the strongest previously reported MultiClinCorpus method. Results are reported using Strict F1, requiring exact agreement of entity type and character offsets. $\Delta$F1 denotes the absolute improvement over the previous best method.}
\label{tab:comparison-best-methods}

\small
\setlength{\tabcolsep}{4.5pt}
\renewcommand{\arraystretch}{1.08}

\begin{tabular}{lllrlrr}
\toprule
Lang & Entity & Previous method & Previous F1 & Best model & Strict F1 & $\Delta$F1 \\
\midrule

\multirow{3}{*}{EN}
& Disease   & CA26AM+MCAI & 0.8960 & GLM 5.2 & \textbf{0.9680} & +0.0720 \\
& Procedure & CA26AM+MCAI & 0.8410 & GLM 5.2 & \textbf{0.9387} & +0.0977 \\
& Symptom   & CA26AM+MCAI & 0.8790 & GLM 5.2 & \textbf{0.9534} & +0.0744 \\

\midrule

\multirow{3}{*}{CS}
& Disease   & CA26AM+MCAI & 0.8510 & GLM 5.2 & \textbf{0.9278} & +0.0768 \\
& Procedure & CA26AM+MCAI & 0.8190 & GLM 5.2 & \textbf{0.8946} & +0.0756 \\
& Symptom   & CA26AM+MCAI & 0.8110 & GLM 5.2 & \textbf{0.9163} & +0.1053 \\

\midrule

\multirow{3}{*}{IT}
& Disease   & CA26AM+MCAI & 0.8820 & Gemma4:31B & \textbf{0.9688} & +0.0868 \\
& Procedure & CA26AM+MCAI & 0.7990 & Gemma4:31B & \textbf{0.9502} & +0.1512 \\
& Symptom   & CA26AM+MCAI & 0.8310 & GLM 5.2 & \textbf{0.9394} & +0.1084 \\

\midrule

\multirow{3}{*}{NL}
& Disease   & CA26AM+MCAI & 0.8200 & GLM 5.2 & \textbf{0.8941} & +0.0741 \\
& Procedure & CA26AM+MCAI & 0.7700 & GLM 5.2 & \textbf{0.8563} & +0.0863 \\
& Symptom   & CA26AM+MCAI & 0.7680 & Gemma4:31B & \textbf{0.8327} & +0.0647 \\

\midrule

\multirow{3}{*}{RO}
& Disease   & CA26AM+MCAI & 0.8980 & GLM 5.2 & \textbf{0.9582} & +0.0602 \\
& Procedure & CA26AM+MCAI & 0.8560 & GLM 5.2 & \textbf{0.9469} & +0.0909 \\
& Symptom   & CA26AM+MCAI & 0.8610 & GLM 5.2 & \textbf{0.9485} & +0.0875 \\

\midrule

\multirow{3}{*}{SV}
& Disease   & CA26AM+MCAI & 0.8280 & GLM 5.2 & \textbf{0.9090} & +0.0810 \\
& Procedure & CA26AM+MCAI & 0.8060 & Gemma4:31B & \textbf{0.8937} & +0.0877 \\
& Symptom   & CA26AM+MCAI & 0.8110 & Gemma4:31B & \textbf{0.8885} & +0.0775 \\
\midrule
\multicolumn{3}{l}{\textbf{Average}} & 0.8348 & -- & \textbf{0.9214} & +0.0866 \\
\bottomrule
\end{tabular}
\end{table*}

The gains observed with document-level projection were preserved under the strictest evaluation criterion, which requires exact agreement of entity type and character boundaries. As shown in Table~\ref{tab:comparison-best-methods}, the best document-level configuration exceeded the strongest previously reported MultiClinCorpus method, CA26AM+MCAI~\cite{remy-2026-parallia}, in all 18 language--entity combinations. Mean Strict F1 increased from 0.8348 to 0.9214, an absolute improvement of 0.0866. Gains were observed for every language and entity type and ranged from +0.0602 for Romanian \textsc{Disease} to +0.1512 for Italian \textsc{Procedure}. The consistency of these improvements under exact-span evaluation is particularly relevant for corpus construction, because successful projection requires not only localisation of the corresponding clinical concept but also recovery of the precise textual boundaries needed to generate reproducible character-level annotations.

\subsection{Accuracy-Efficiency Trade-offs}
\label{sec:results-efficiency}

The projection paradigms exhibited substantially different computational profiles. GLM 5.2 required an average of 6.54 s per document through cloud inference, whereas locally deployed Gemma4:31B required 27.01 s per document, with a mean generation rate of 33.32 tokens/s. In comparison, our local open-source implementation of the ClinicalAligner\footnote{\href{https://github.com/alvaroreyb/cross-lingual-biomedical-aligner}{BiomedicalAligner}} approach required approximately 0.1 s per document after training.

ClinicalAligner additionally required a dedicated training stage, estimated at approximately 2--4 h on an RTX 3090 in our implementation, whereas the LLM approaches can be applied directly without task-specific training. Under these measurements, the approximate point at which the initial alignment-model training cost is amortised is 1,118--2,236 documents relative to GLM 5.2 and 268--535 documents relative to Gemma4:31B. These values should be interpreted as implementation- and hardware-dependent estimates rather than intrinsic properties of the models.

\section{Discussion}
\label{sec:discussion}

Our findings indicate that cross-lingual clinical annotation projection can be effectively reformulated as a constrained document-level generation problem, provided that generative inference is coupled with deterministic text-preservation validation and character-level grounding. This formulation substantially outperformed the current state of the art for cross-lingual clinical annotation projection. Rather than relying on explicit alignment or candidate generation and ranking, the LLM operates on the complete parallel documents and directly reconstructs the annotated target text. This formulation appears particularly well suited to clinical projection, where translation often preserves the underlying clinical information while modifying lexical form, word order, or entity boundaries.

Gemma4:31B is particularly relevant because, like CA26AM+MCAI, it can be deployed entirely locally. Its consistent advantage over the previous state of the art despite requiring no specialised alignment architecture indicates that document-level generation can improve both semantic localisation and exact span reconstruction. The strict improvement ranged from +0.0564 to +0.1512 F1, with the largest gains observed for \textsc{Procedure} entities. These gains indicate not only successful semantic localisation but also more accurate reconstruction of the complete annotated span.

The results across languages also show that projection difficulty cannot be explained solely by the expected availability of language resources. English, Italian, and Romanian generally produce the strongest LLM results, whereas Dutch is the most challenging target language. Czech and Swedish, despite being comparatively less represented in many multilingual NLP resources, remain competitive under direct LLM projection. This suggests that document-level contextual reasoning can partially compensate for weaker lexical correspondence or lower representation during multilingual pretraining, while also indicating that linguistic resource availability alone does not determine projection performance.

A clearer linguistic pattern emerges for the ML approach. Among the non-English targets, Italian and Romanian benefit most from candidate-based projection, while performance decreases for more distant languages and is particularly limited for Czech. This behaviour is consistent with the feature representation used by the classifier, which combines lexical overlap, edit similarity, character and word $n$-grams, positional information, and multilingual semantic similarity. Romance languages provide stronger surface and structural correspondence with Spanish, making the candidate-ranking problem easier. In contrast, direct LLM projection exhibits substantially less dependence on these explicit form-level similarities.

The results obtained with the hybrid strategy further suggest that the main advantage of LLM-based projection arises when the model can reason over the complete projection problem rather than being restricted to revising candidates produced by an upstream system. Candidate-based methods remain attractive because they substantially constrain the search space and reduce inference requirements, but their attainable performance is limited by candidate generation: if the correct target span is absent or poorly represented among the candidates, subsequent classification or generative refinement cannot fully recover it. Direct projection removes this dependency by locating and annotating the corresponding entity directly within the complete target document.

These findings are particularly relevant for multilingual corpus construction. A single LLM can be applied across languages and entity types without training separate projection classifiers or maintaining a specialised alignment model for each setting. In particular, locally deployable models make it possible to retain the complete projection workflow within institutional infrastructure, an important consideration in clinical NLP settings in which governance, privacy, or data-sharing restrictions may preclude externally hosted inference services.

Nevertheless, direct LLM projection introduces failure modes that differ from those of conventional alignment systems. Generative models may alter the target text, omit annotations, duplicate entities, or produce malformed tags. Deterministic validation and offset reconstruction are therefore not merely post-processing conveniences but integral components of the proposed formulation: they ensure that accepted projections preserve the original target text and can be converted into reproducible character-level annotations before incorporation into the resulting corpus.

The different projection paradigms also involve a clear accuracy--efficiency trade-off. Once training has been amortised, alignment-based projection remains considerably more computationally efficient for large or repeatedly processed collections. Direct LLM projection instead exchanges higher per-document inference cost for substantially higher exact-span accuracy, the absence of task-specific training, and a simpler projection pipeline. This trade-off is particularly relevant for local deployment, where both LLM-based and alignment-based approaches can operate without external inference services and the preferred strategy can therefore be selected according to corpus size, computational resources, and required annotation quality.

\subsection{Limitations}
\label{sec:limitations}

This study has several limitations. First, all experiments were conducted within the MultiClinCorpus benchmark, using Spanish as the single source language and six European target languages. Although these languages include Romance, Germanic, and Slavic families, the extent to which the observed robustness generalises to other source languages, more typologically distant target languages, or independently authored multilingual clinical documents remains unknown. The evaluation was also restricted to three entity types---\textsc{Disease}, \textsc{Symptom}, and \textsc{Procedure}---and therefore does not establish performance for other clinical concepts or more complex annotation structures. External validation on additional clinical corpora and language pairs will be necessary to assess the generalisability of the proposed formulation.

Second, evaluation relied on the official gold-standard span annotations and quantified exact and partial boundary agreement, but we did not conduct an additional bilingual or clinical-expert error analysis of the residual projections. Deterministic validation verifies tag structure, entity counts, text preservation, and character-offset reconstruction, but cannot by itself establish the semantic correctness of a structurally valid projected span. Similarly, the present study evaluates annotation projection directly rather than the downstream utility of the resulting projected corpora; future work should determine whether models trained on automatically projected annotations retain performance when applied to independently annotated, native-language clinical text. Finally, only three LLMs and specific inference configurations were evaluated, and computational measurements depend on model implementation, serving infrastructure, and hardware. Accordingly, absolute efficiency estimates and the relative advantage of particular models should not be interpreted as invariant across deployment environments or future model versions.

\section{Conclusion}
\label{sec:conclusion}

This study demonstrates that cross-lingual clinical annotation projection can be formulated as a constrained, text-preserving document-level generative task capable of producing accurate and verifiable character-level annotations across multiple languages. Direct LLM projection consistently outperformed candidate-based supervised, hybrid, and previous state-of-the-art projection approaches across six target languages and three clinical entity types, while avoiding the need for explicit alignment, candidate generation, or task-specific training.

Across the evaluated corpus, the best LLM configuration for each language--entity combination produced 55,416 projected clinical mentions across six target languages and three entity types. By transferring expert annotations from a single source language while preserving exact textual grounding and reproducible character offsets, the proposed formulation provides a practical mechanism for extending annotated clinical resources, particularly where manually annotated corpora and language-specific NLP resources remain limited.

Although direct LLM projection entails higher inference costs than specialised alignment-based approaches, it combines strong exact-span accuracy with a simpler task-specific workflow and, when locally deployed, can operate entirely within institutional infrastructure. Coupled with deterministic validation and offset reconstruction, this makes text-preserving document-level projection a practical and auditable strategy for multilingual clinical corpus construction, with the potential to reduce the amount of expert annotation required when extending existing resources across languages.

\section*{Funding}
This work was supported by the Ministerio de Ciencia e Innovación (MICINN) under project PID2024-155334OB-I00. Funding for the open access charge was provided by Universidad de Málaga / CBUA.

\bibliographystyle{plain}
\bibliography{references}
\clearpage
\appendix
\renewcommand{\thelstlisting}{A\arabic{lstlisting}}
\setcounter{lstlisting}{0}
\section{Prompts}
\label{app:promps}

\begin{lstlisting}[style=prompt, caption={Match-classification prompt in English}, label={lst:annot-prompt}, literate={á}{{\'a}}1 {é}{{\'e}}1 {í}{{\'i}}1 {ó}{{\'o}}1 {ú}{{\'u}}1 {ñ}{{\~n}}1 {ü}{{\"u}}1]
You are a strict bilingual clinical text matcher.

Your task is to determine whether a <SOURCE_LANGUAGE> entity mention is represented in a <TARGET_LANGUAGE> text window.
If there is any noise, consider it a no match.

Source entity:
<SOURCE_ENTITY>

Target window:
<TARGET_WINDOW>

Return a JSON object with this exact schema:
{
  "match_label": "exact match | mid match | no match",
  "justification": "string"
}

Label definitions:
- exact match: the target window clearly contains the same concept or a direct translation of the source entity
- mid match: partial, approximate, broader, narrower, abbreviated, or contextually related match
- no match: the target window does not represent the same concept

Rules:
- Compare meaning, not literal wording
- Consider translations, abbreviations, paraphrases, and morphological variants
- If uncertain between exact match and mid match, choose mid match
- If match_label is "exact match", justification must be an empty string
- If match_label is "mid match" or "no match", justification must contain a brief explanation
- Return JSON only

\end{lstlisting}
\begin{lstlisting}[style=prompt, caption={Span correction prompt in English}, label={lst:annot-prompt}, literate={á}{{\'a}}1 {é}{{\'e}}1 {í}{{\'i}}1 {ó}{{\'o}}1 {ú}{{\'u}}1 {ñ}{{\~n}}1 {ü}{{\"u}}1]


You are a biomedical NER expert.
Below is a clinical document in <TARGET_LANGUAGE>, followed by entity spans that were not good matches.
For each entity, use the document context and the <SOURCE_LANGUAGE> reference to suggest the most accurate <TARGET_LANGUAGE> entity string from the document.

DOCUMENT:
<TARGET_DOCUMENT>

PROBLEMATIC ENTITIES:
<PROBLEMATIC_ENTITIES>

Respond ONLY with a JSON array, no explanation:
[{"id": 1, "suggestion": "..."}, ...]
\end{lstlisting}

\begin{lstlisting}[style=prompt, caption={Cross-lingual projection prompt in Spanish}, label={lst:annot-prompt}, literate={á}{{\'a}}1 {é}{{\'e}}1 {í}{{\'i}}1 {ó}{{\'o}}1 {ú}{{\'u}}1 {ñ}{{\~n}}1 {ü}{{\"u}}1]
Eres especialista en anotación de corpus clínicos multilingües. Proyecta únicamente las siguientes etiquetas XML desde el documento fuente en español al documento traducido. Conserva exactamente el texto traducido y devuelve únicamente el documento final en el idioma que corresponda con las etiquetas insertadas.

Proyecta al documento traducido las etiquetas XML clínicas presentes en el documento español.

Reglas obligatorias:

Usa únicamente estas etiquetas: {ACTIVE_LABELS}.
Conserva exactamente el texto traducido; solo puedes insertar etiquetas XML.
Mantén el mismo número de etiquetas de apertura y cierre para cada etiqueta que en el documento español.
No añadas explicaciones, markdown, comillas ni ningún texto adicional.
Devuelve únicamente el documento traducido etiquetado.

Documento español etiquetado:
<documento_español_etiquetado_xml>
{SPANISH_XML_DOCUMENT}
</documento_español_etiquetado_xml>

Documento traducido sin etiquetas:
<documento_traducido>
{TARGET_DOCUMENT}
</documento_traducido>

\end{lstlisting}

\begin{lstlisting}[style=prompt, caption={Cross-lingual projection prompt in English}, label={lst:annot-prompt}, literate={á}{{\'a}}1 {é}{{\'e}}1 {í}{{\'i}}1 {ó}{{\'o}}1 {ú}{{\'u}}1 {ñ}{{\~n}}1 {ü}{{\"u}}1]
You are a specialist in multilingual clinical corpus annotation. Project only the following XML tags from the Spanish source document into the translated document. Preserve the translated text exactly and return only the final document in the appropriate language with the tags inserted.

Project the clinical XML tags present in the Spanish document onto the translated document.

Mandatory rules:

Use only these tags: {ACTIVE_LABELS}.
Preserve the translated text exactly; you may only insert XML tags.
Maintain the same number of opening and closing tags for each label as in the Spanish document.
Do not add explanations, markdown, quotation marks, or any additional text.
Return only the tagged translated document.

Tagged Spanish document:
<spanish_doc_tagged_xml>
{SPANISH_XML_DOCUMENT}
</spanish_doc_tagged_xml>

Untagged translated document:
<translated_document>
{TARGET_DOCUMENT}
</translated_document>


\end{lstlisting}

\end{document}